\documentclass[12pt]{article}
\usepackage{amstext,amssymb,amsmath,epsf,graphics}

\renewcommand{\geq}{\geqslant}
\renewcommand{\leq}{\leqslant}

\newcommand{\bm}[1]{\boldsymbol #1}
\newcommand{\Real}{\mbox{$\mathbb{R}$}}

\newcommand{\define}{\mbox{$\triangleq$}}

\begin{document}

\title{Stretchy Polynomial Regression}

\author{Kar-Ann Toh\\ School of EEE, Yonsei University\\ katoh@ieee.org}
\date{August 2014}

\maketitle

\begin{abstract}
This article proposes a novel solution for stretchy polynomial regression
learning. The solution comes in primal and dual closed-forms similar to that of
ridge regression. Essentially, the proposed solution stretches the covariance
computation via a power term thereby compresses or amplifies the estimation.
Our experiments on both synthetic data and real-world data show effectiveness
of the proposed method for compressive learning.
\end{abstract}

\section{Introduction}

The Weierstrass's approximation theory (see e.g., \cite{WadeW1}) states that
polynomials can approximate any continuous function on a closed and bounded
interval to any degree of accuracy. This means that multivariate polynomials
can provide an effective way to describe complex nonlinear input-output
relationships \cite{Toh39}.

However, on top of the commonly encountered heavy computational requirement,
the large number of polynomial expansion terms arising from high dimensional
systems and high model orders often gives rise to an under-determined or
over-complete system when the number of training samples is small. These are
the main reasons that full multivariate polynomials, particularly beyond third
orders, are seldom adopted in real world applications.

In this article, we attempt to handle the resulting under-determined or
over-complete systems through coefficient shrinkage. Two novel solutions in
primal and dual closed-forms are proposed to stretch the regression beyond
existing frameworks. Since the proposed solutions work only on positive real
input space, an exponential transformation is proposed to convert standardized
inputs to the first quadrant of real axis. Attributed to the additional degree
of freedom in twisting the input space, this transformation provides a
mechanism to further stretch the above regression for possible compressive
learning.

Our contributions of this work include: (i) proposal of a smooth and
closed-form \emph{stretchy regression} for compressive learning; (ii) proposal
of an input transformation to further stretch possible compressive learning.
(iii) illustration of possible use of full multivariate polynomials for high
model orders for regression applications.

\section{Preliminaries}

\subsection{Linear Models}

Linear estimation models are among the most popular choices for data fitting
and they remained to be among our most important tools. Given a set of training
data which consists of $M$ examples $(\bm{x}_i,y_i)$, $i=1,...,M$, where
$\bm{x}_i\in\Real^d$ denotes the $i^{th}$ feature sample, and $y_i\in\Real$
denotes the corresponding \emph{target output}. In other words, the value $y_i$
can be viewed as the output associated with $\bm{x}_i$ in the system to be
learned. Using the given feature sample as input, a \emph{predictor} outputs a
value which can be associated with target prediction.

In single-output regression, the goal is to determine a predictor $g(\cdot)$ to
fit the target output $y$ (with sample index omitted here). In binary
classification, $y\in\{0,1\}$ or $y\in\{-1,+1\}$, the goal is to determine a
predictor $g(\cdot)$ plus a threshold value $\tau$ such that a correct class
prediction can be obtained for unseen data. An ideal classifier is such that
$cls(g(\bm{x}_j)\geq\tau) = y_j$ for all unseen samples indexed by $j=1,2,
\ldots, N$ where $cls(\cdot)$ denotes a classification function which outputs
either $\{0,1\}$ or $\{-1,+1\}$ based on the decision threshold $\tau$.

Typically for a single data sample with its sample index omitted, a linear
predictor model can be written as
\begin{equation}\label{eqn_linear_model}
    g(\bm{x},\bm{\alpha}) = \alpha_0 + \sum^d_{j=1} x_j \alpha_j = \bm{x}^T\bm{\alpha},
\end{equation}
where the notation of the right most expression has the \emph{intercept} or
\emph{bias} term being absorbed into the vector expression giving
$\bm{\alpha}=[\alpha_0,\alpha_1,...,\alpha_d]^T$ and
$\bm{x}=[1,x_1,...,x_d]^T$. A generalized linear model \cite{Duda1} can have
its inputs expanded to a transformed space giving
\begin{equation}\label{eqn_linear_model_p}
    g(\bm{x},\bm{\alpha}) = \alpha_0 + \sum^{D}_{j=1} p_j(\bm{x}) \alpha_j = \bm{p}^T\bm{\alpha},
\end{equation}
where $\bm{p}$ transforms $\bm{x}$ from $\Real^{d+1}$ to $\Real^{D+1}$, and
$\bm{\alpha}\in\Real^{D+1}$. The variate $\bm{p}$ is also called a \emph{basis}
expansion term with popular choice taking the form of \emph{gaussian} or
\emph{sigmoid} function. Our proposed polynomial expansion falls within this
generalized linear model form where its linearity is with respect to the
parameter vector $\bm{\alpha}$.

For multiple data samples, the arising multiple column vectors of $\bm{p}$ can
be stacked as ${\bf P}=[\bm{p}_1,...,\bm{p}_M]^T$ where the generalized linear
model \cite{Toh77} can be compactly written as
\begin{equation}\label{eqn_linear_model_BigP}
    {\bf g}(\bm{x},\bm{\alpha}) = {\bf P}\bm{\alpha}.
\end{equation}

\subsection{Full Multivariate Polynomials} \label{sec_FMP}

A general multivariate polynomial model of order $r$ can be expressed as
\begin{equation}
  g(\bm{\alpha},\bm{x}) = \sum_i  \alpha_i x_1^{n_1}x_2^{n_2}\cdots x_d^{n_d},
  \label{eqn_full_MP}
\end{equation}
where the summation is taken over all non-negative integers $n_1,n_2,...,n_d$
for which $n_1+n_2+\cdots+n_d\leq r$. The total number of terms in
$g(\bm{\alpha},\bm{x})$ is given by $\displaystyle{D = \sum_{k=0}^r \
\frac{(k+d-1)! }{ k! (d-1)!}}$ where $0!=1$. The parameter vector
$\bm{\alpha}=[\alpha_0,\alpha_1,...,\alpha_D]^T$ is
 to be estimated, while the input regressor vector  $\bm{x}= [1,x_1,...,x_d]^T$ contains $d$ input
 features ($d$-dimensional input) with an intercept term. Without loss of generality, we assume
 the input is normalized such that $x_j\in(0,1)$, $j=1,...,d$. This inherently
 implies that all polynomial product terms are also bounded within the
 unit interval.

\subsection{Compressive Learning}

Suppose $p\geq 1$ is a real number. The commonly known $p$-norm for parameter
vector $\bm{\alpha}$ is defined as
\begin{equation}\label{eqn_p_norm}
    \ell^p:\hspace{5mm} \|\bm{\alpha}\|_p \define \left(\sum^D_{i=0}|\alpha_i|^p \right)^{1/p}.
\end{equation}
When $p=1$, it is commonly known as $\ell^1$-norm or \emph{taxicab}-norm. For
$p=2$, we have the well-known Euclidean norm and when $p$ approaches infinity,
the $p$-norm approaches the \emph{infinity}-norm or \emph{maximum}-norm.
However, for $0<p<1$, the resulting function does not define a norm since the
\emph{triangle inequality} is violated. Nevertheless, it remains true that the
function $\int_X |f(x)-g(x)|^p d\mu$ defines a \emph{distance} which makes
$\ell^p(X)$ a complete metric \emph{topological vector space}.

\begin{figure}[hhh]
  \begin{center}
  \epsfxsize=8.8cm
  \epsfysize=5.8cm
  \epsffile[58   195   546   602]{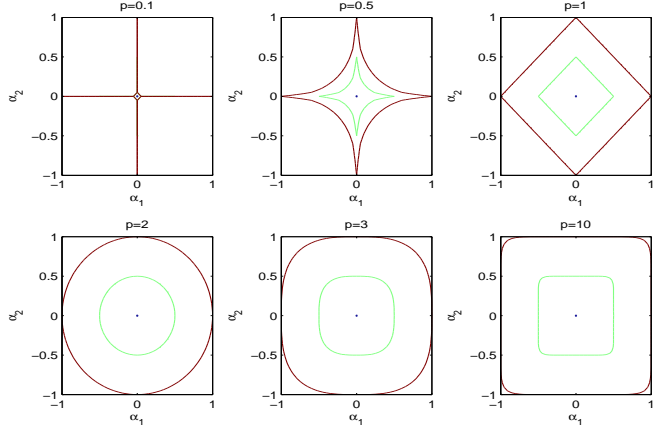}
  \caption{A two-dimensional $p$-space for $p\in\{0.1,0.5,1,2,3,10\}$}
  \label{fig_p_norm}
  \end{center}
\end{figure}

In least squares related regularization and coefficient shrinkage, the
following $p$ values are of particular interest:
\begin{itemize}
    \item $p=2$ \cite{Hoerl1,Hoerl2}: This is called \emph{ridge regression} where a stable but dense
                 estimation solution is obtained.
    \item $p=1$ \cite{Tibshirani1}: This is called \emph{lasso} where a moderately sparse estimation solution
                can often be obtained.
    \item $p=0$ \cite{Miller1}: This is termed \emph{subsets selection} where the sparest
                 estimation solution is inferred.
    \item $0\leq p\leq 2$ \cite{Frank1}: This is called \emph{bridge regression} which bridges between subset selection and ridge
    regression.
    \item $1\leq p\leq 2$ \cite{Zou1}: This is called \emph{elastic net} which bridges between lasso and ridge
    regression.
\end{itemize}
Here we note that $p\geq 1$ implies convexity while $p<1$ implies non-convexity
in the solution space. Fig.~\ref{fig_p_norm} shows the contour plots within a
unit ``cube'' of estimation solution space for $\bm{\alpha}$ in two-dimension
for various $p$-values.

\section{Proposed Stretchy Regression}

\subsection{Coefficient Shrinkage}

\begin{figure}[hhh]
  \begin{center}
  \epsfxsize=10.8cm
  \epsfysize=16.8cm
  \epsffile[58   208   546   594]{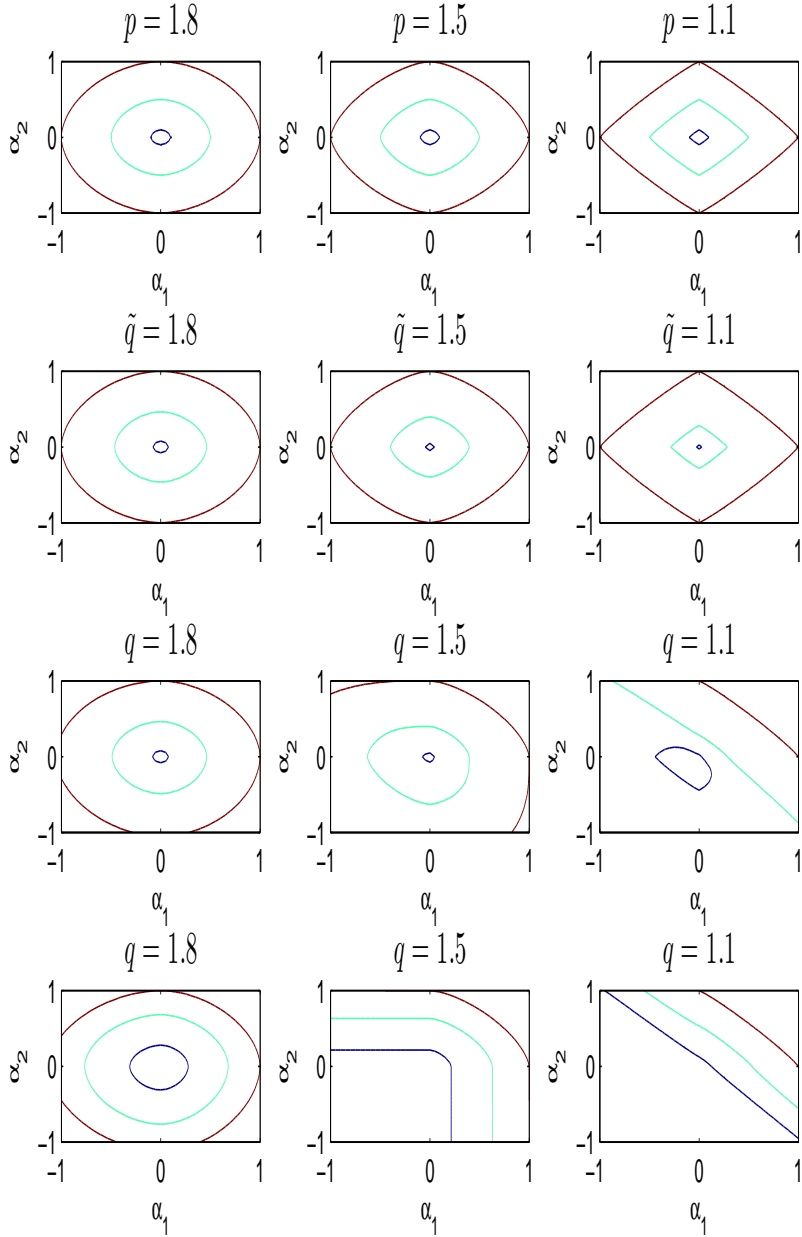}
  \caption{Contour plots at levels 0.1, 0.5 and 1. Top panel: $p$-norms for $p\in\{1.8,1.5,1.1\}$,
  Second Panel: $\tilde{q}$-space for $q\in\{1.8,1.5,1.1\}$,
  Third Panel: real $q$-space for $q\in\{1.8,1.5,1.1\}$,
  Bottom Panel: $q^2$-space for $q\in\{1.8,1.5,1.1\}$. }
  \label{fig_q_norm}
  \end{center}
\end{figure}

Consider a real integer $q$ defined on the following modified space (called
$\tilde{q}$-space for convenience):
\begin{equation}\label{eqn_q_norm}
     \|\bm{\alpha}\|_{\tilde{q}} \define \left(\sum^D_{i=0}|\alpha_i|^{q} \right)^{1/2},
\end{equation}
where $q\in\Real$. When the absolute operator for $\alpha_i$ is omitted, we
have a modified form for \eqref{eqn_q_norm} (somewhat related to the
\emph{generalized mean} without averaging) as follows:
\begin{equation}\label{eqn_q_norm_approx}
    \|\bm{\alpha}\|_{q} \define \left(\sum^D_{i=0}\alpha_i^{q}
    \right)^{1/2}.
\end{equation}
In a loose sense, we shall call \eqref{eqn_q_norm_approx} a \emph{$q$-space}
for convenience hereon (notice that this is not a \emph{normed} vector space
since the scaling property is violated). Fig.~\ref{fig_q_norm} shows the
$\ell^{p}$-space \eqref{eqn_p_norm} for $1<p<2$ and the corresponding
$\tilde{q}$ \eqref{eqn_q_norm}, $q$ \eqref{eqn_q_norm_approx} and $q^2$-spaces
within the same interval. Here we see that the plots for $\tilde{q}$-space show
much resemblance to those of $\ell^{p}$-space \eqref{eqn_p_norm}. From the
bottom two panels of Fig.~\ref{fig_q_norm}, we see that the positive quadrant
of the solution of \emph{real} $q$-space and $q^2$-space fits well to the
solution $p$-space for $1<p<2$. This suggests vertices with positive values
being feasible solutions for the proposed constrained solution space. This
observation shall be exploited in the following development.

\clearpage

Next, consider the following minimization problem:
\begin{equation}\label{eqn_q_norm_obj111}
    \min_{\bm{\alpha}} \|\bm{\alpha}\|^2_{q}
    + \bm{\beta}^T({\bf y}-{\bf P}\bm{\alpha}).
\end{equation}
Denote the \emph{elementwise} Hadamard product between vectors
$\bm{a}\in\Real^d$ and $\bm{b}\in\Real^d$ as $\bm{a}\circ\bm{b}$. Also, in
order to simplify notations, all the \emph{elementwise} power terms of vector
and matrix in what follows shall be denoted as $\bm{a}^q$ or ${\bf A}^q$ except
for inverse of square matrix. Let
$\bar{\bm{\alpha}}\define[\alpha_1^{q/2},\cdots,\alpha_d^{q/2}]=\bm{\alpha}^{q/2}$
where we can write
$\|\bm{\alpha}\|_{q}=(\bar{\bm{\alpha}}^T\bar{\bm{\alpha}})^{1/2}$. Then take
the first derivative of \eqref{eqn_q_norm_obj111} and set it to zero gives:
\begin{eqnarray}
   \bm{\alpha} & = & \left(\frac{1}{q}{\bf P}^T\bm{\beta}\right)^{\frac{1}{q-1}}. \label{eqn_qnorm_alpha111}
\end{eqnarray}
Based on Newton's generalized binomial theorem, we consider a scaling vector
$\bm{s}$ which factors out ${\bf P}$ and $\bm{\beta}$ in the following manner:
\begin{eqnarray}\label{eqn_scaling_factor}
    \left({\bf P}^T\bm{\beta}\right)^{\frac{1}{q-1}}  &=& (
    {\bf P}^T)^{\frac{1}{q-1}}\bm{\beta}^{\frac{1}{q-1}}\circ\bm{s}.
\end{eqnarray}
Then multiply both sides of \eqref{eqn_qnorm_alpha111} by ${\bf P}$ and replace
${\bf P}\bm{\alpha}$ by ${\bf y}$ gives
\begin{eqnarray}
   \bm{\beta} &=& q\left[\left({\bf P}({\bf P}^T)^{\frac{1}{q-1}}\right)^{-1}{\bf y}\circ\bm{s}^{-1}
   \right]^{q-1} .
\end{eqnarray}
Here we note that only the term $\left({\bf P}({\bf
P}^T)^{\frac{1}{q-1}}\right)^{-1}$ involves a full matrix inverse while all
other power terms are elementwise operation. Substitute $\bm{\beta}$ into
\eqref{eqn_qnorm_alpha111} and simplify gives:
\begin{eqnarray}
  \bm{\alpha}
   &=& \left({\bf P}^T\right)^{\frac{1}{q-1}}\left[{\bf P}
        ({\bf P}^T)^{\frac{1}{q-1}}\right]^{-1}{\bf y} . \label{eqn_qnorm_dual_soln}
\end{eqnarray}
Notice that apart from the power terms, this solution form is analogous to that
of dual ridge regression.

Next, we proceed to convert the above solution in dual space form to its primal
form. Based on the matrix identity $({\bf I}+{\bf A}{\bf B})^{-1}{\bf A}={\bf
A}({\bf I}+{\bf B}{\bf A})^{-1}$, the solution \eqref{eqn_qnorm_dual_soln}
under dual space can thus be re-written in primal space as
\begin{equation}\label{eqn_qnorm_primal_soln}
  \bm{\alpha} = \left[({\bf P}^T)^{\frac{1}{q-1}}{\bf P}\right]^{-1}
        \left({\bf P}^T\right)^{\frac{1}{q-1}}{\bf y}.
\end{equation}
For data that results in near singularity of the stretched covariances $({\bf
P}^T)^{\frac{1}{q-1}}{\bf P}$ or ${\bf P}({\bf P}^T)^{\frac{1}{q-1}}$, a
regularization term can be included within the inverse term.

In the following experiments, we shall adopt \eqref{eqn_qnorm_dual_soln} when
dealing with under-determined systems and adopt \eqref{eqn_qnorm_primal_soln}
when dealing with over-determined systems.

\subsection{First Quadrant Transformation}

Consider a stacked set of raw training input data given by
\begin{equation}\label{eqn_X}
    {\bf X}_{raw} = \stackrel{
\left[
\begin{array}{cccc}
  {\bf x}_1 & {\bf x}_2 & \cdots & {\bf x}_d \\
\end{array}
\right]
    }{\overbrace{\left[
\begin{array}{cccc}
  x_{11} & x_{12} & \cdots & x_{1d} \\
  x_{21} & x_{22} & \cdots & x_{2d} \\
  \vdots & \vdots & \ddots & \vdots \\
  x_{M1} & x_{M2} & \cdots & x_{Md} \\
\end{array}
\right] }} = \left[
\begin{array}{c}
  \bm{x}_1 \\
  \bm{x}_2 \\
  \vdots \\
  \bm{x}_M \\
\end{array}
\right].
\end{equation}
A standardization is first performed for each data column by a $z$-score
normalization based on the statistics of training set (mean $\mu_{{\bf x}_k}$
and variance $\sigma_{{\bf x}_k}$):
\begin{equation}\label{eqn_zscore}
    \underline{{\bf x}}_k=({\bf x}_k - \mu_{{\bf x}_k})/\sigma_{{\bf x}_k},
    \ k=1,...,d.
\end{equation}
Then, an exponential function is adopted to map the standardized data into the
first quadrant:
\begin{equation}\label{eqn_exponential_transform}
    \breve{{\bf x}}=\exp(a\underline{{\bf x}}_k + \bm{b}_k).
\end{equation}

Here, we note that the exponential transformation serves two purposes: first
quadrant transformation and data warping. The main reason for first quadrant
transformation is to handle the power term ($q$) smaller than 2 where complex
number arises in the proposed \emph{stretchy regression}. We call this a key
absolute space transformation. The data warping mechanism \emph{twists} the
original data such that large values are differentiated far more than small
values or vice versa. This twisting further stretches (or compresses) the
relative difference among the input variables on top of the stretchy
regression.

\section{Synthetic Data}

\begin{figure}[hhh]
  \begin{center}
\begin{tabular}{cc}
  \epsfxsize=6.13cm
  \hspace{-2mm}\epsffile[78   212   552   595]{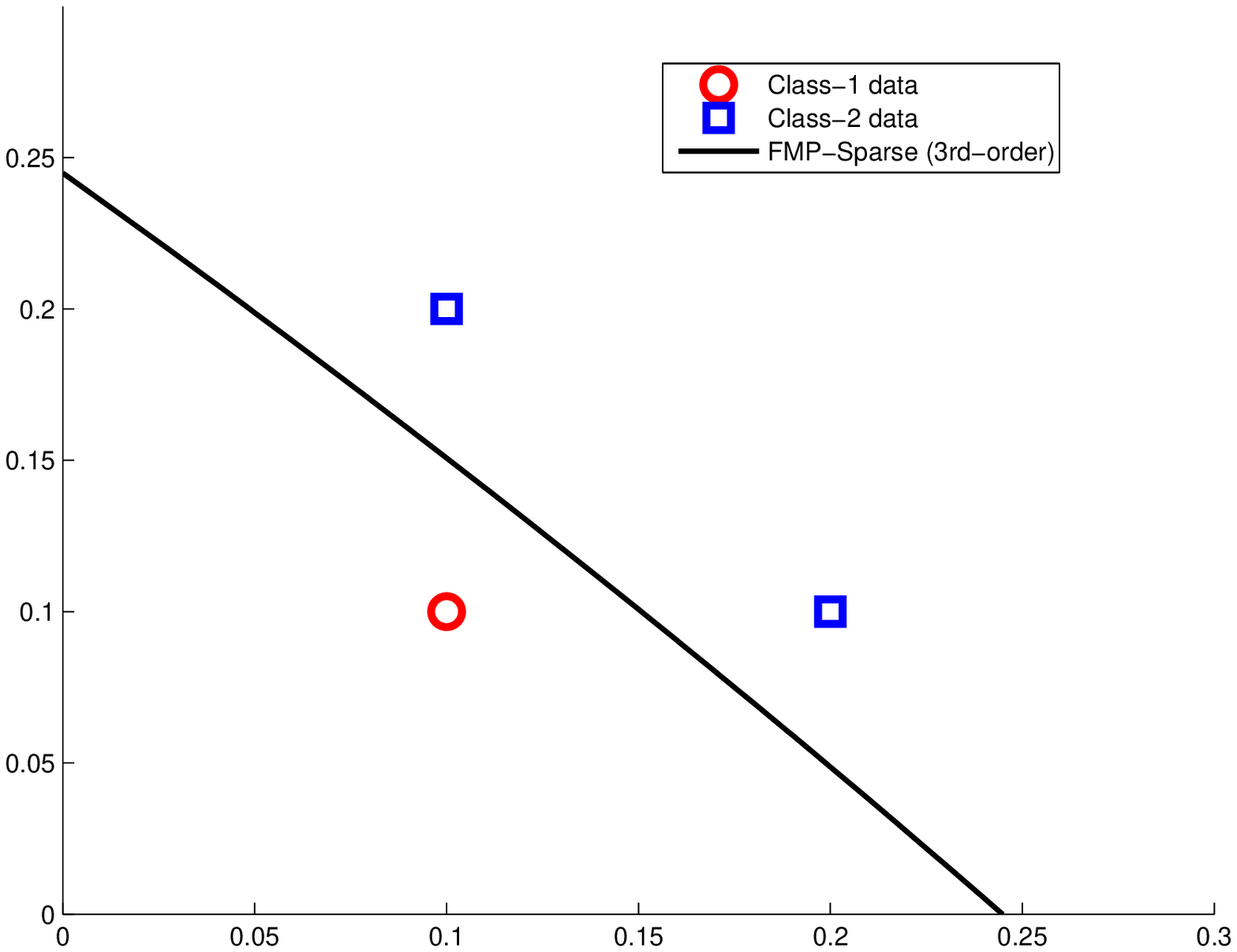} &
  \epsfxsize=6.13cm
  \hspace{-2mm}\epsffile[78   212   552   595]{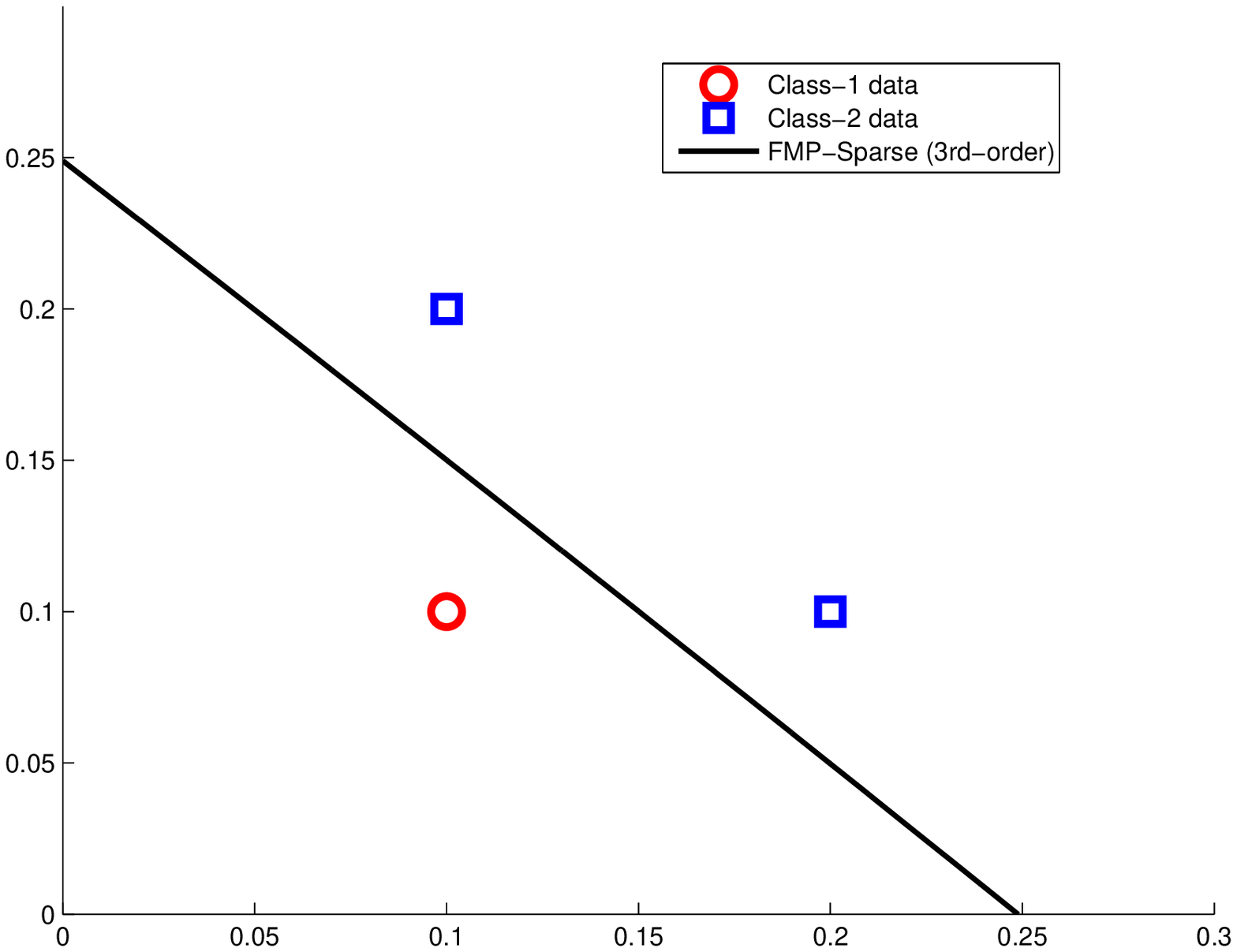}
  \\*[-1mm]  (a) $q=2$  & (b) $q=1.5$ \\*[1mm]
  \epsfxsize=6.13cm
  \hspace{-2mm}\epsffile[78   212   552   595]{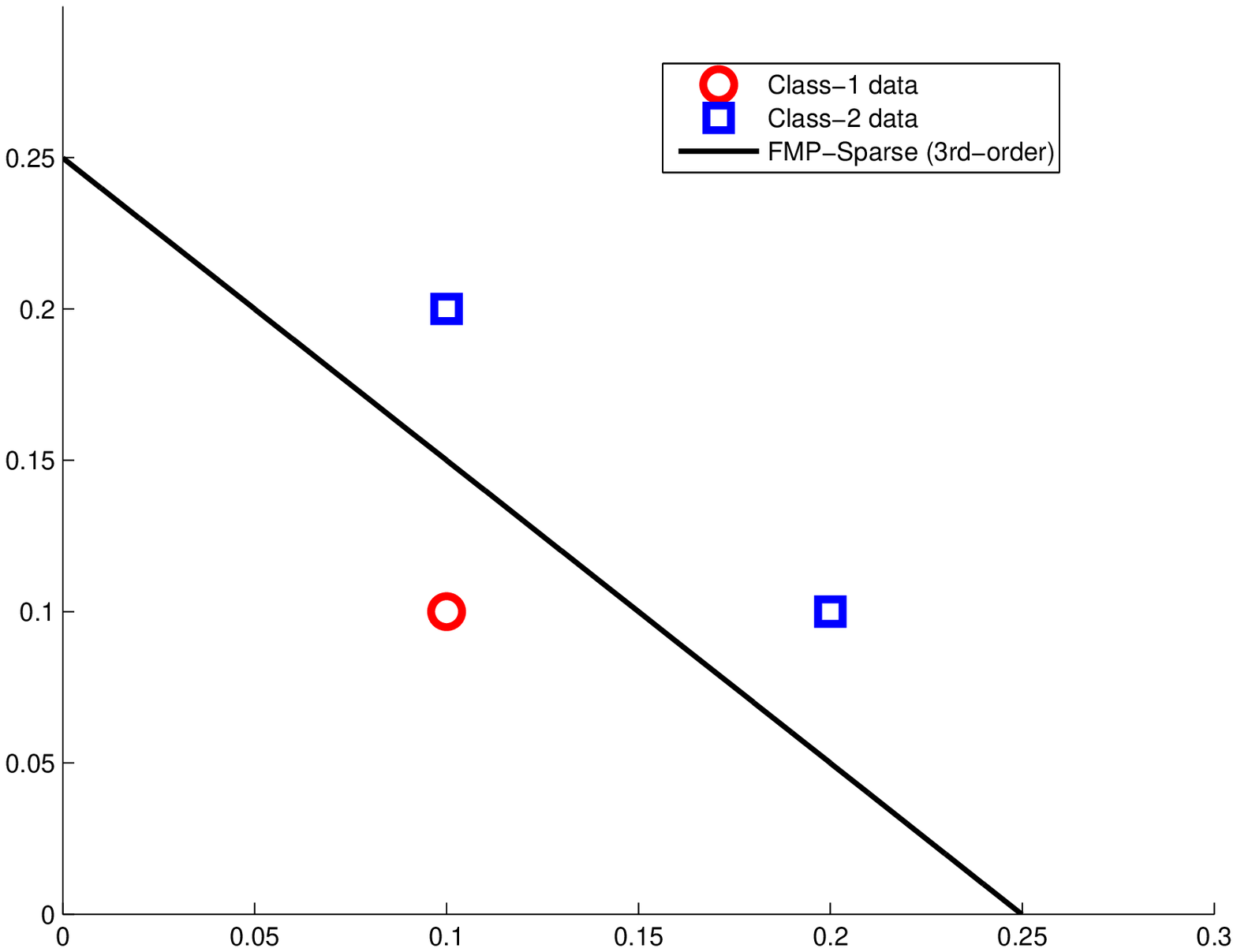} &
  \epsfxsize=6.13cm
  \hspace{-2mm}\epsffile[78   212   552   595]{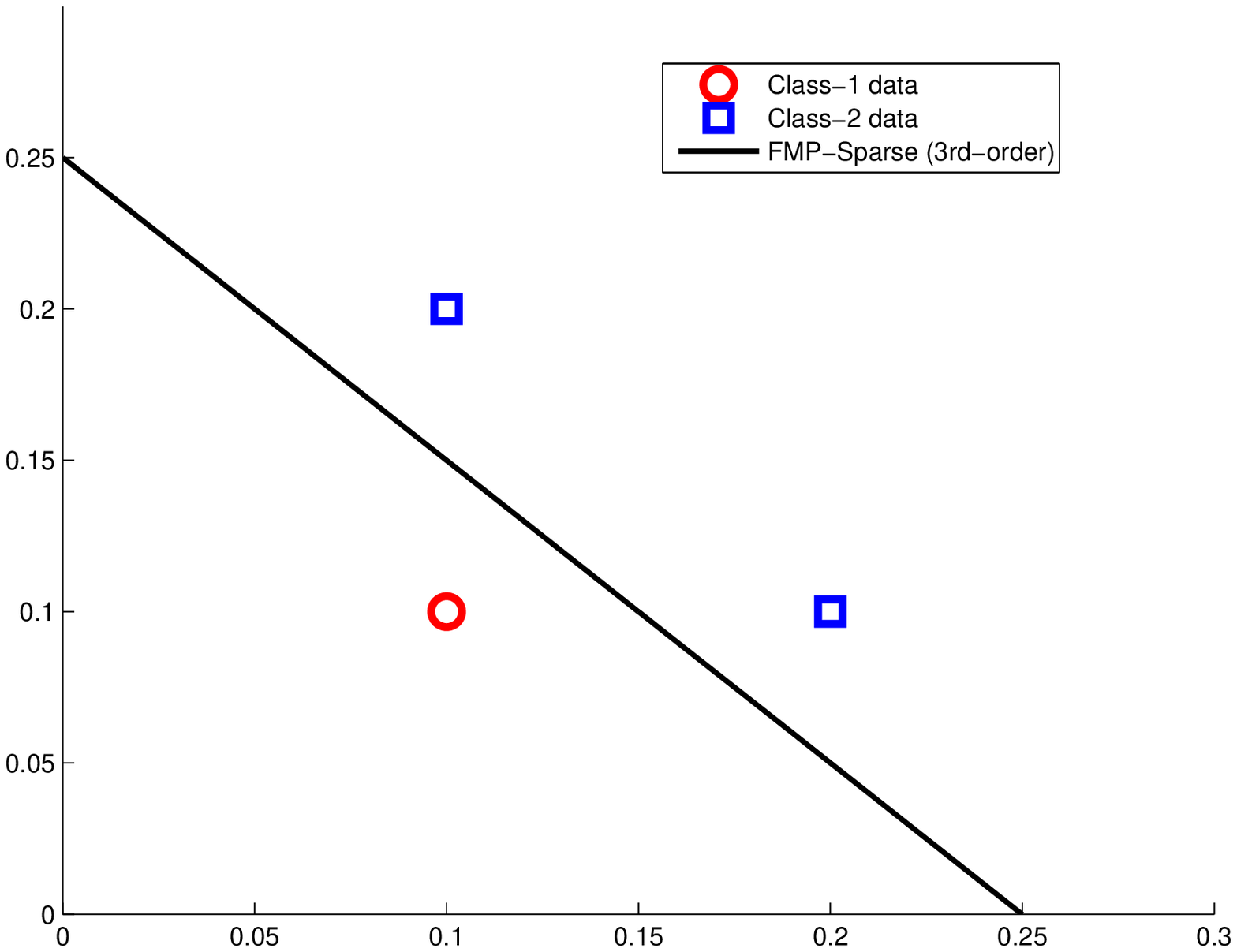}
  \\*[-1mm]  (c) $q=1.3$  & (d) $q=1.1$
\end{tabular}
  \caption{(a) Decision boundaries (zero threshold value) of a 3rd-order FMP-Sparse model
  learned from 3 non-overlapping data points at different $q$ values}  \vspace{-3mm}
  \label{fig_contours1}
  \end{center}
\end{figure}

Consider an example with three synthetic data samples as shown in
Fig.~\ref{fig_contours1} where the red circle indicates a sample drawn from
class-1 distribution ($y=-1$) and the two blue boxes are samples drawn from
class-2 distribution ($y=+1$). Since these data are already in first quadrant,
no standardization and transformation is necessary. The four sub-figures show
the corresponding decision boundaries obtained for a 3rd-order polynomial model
at four $q$ values using \eqref{eqn_qnorm_dual_soln}. These results show
convergence of learning solutions even though the system is under-determined.

\begin{figure}[hhh]
  \begin{center}
  \epsfxsize=12cm
  \epsfysize=8cm
  \epsffile[52   194   555   589]{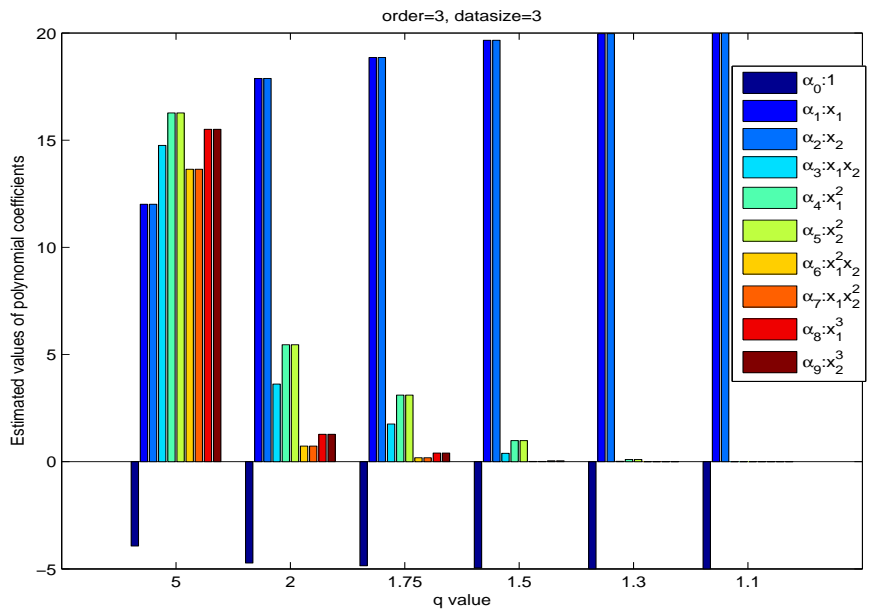}
  \caption{Estimated coefficient values versus $q$ value ($q\in\{2,1.75,1.5,1.3,1.1\}$)
  based on the 3-samples data}
  \label{fig_bargraph_coeff_sparse}
  \end{center}
\end{figure}

Table~\ref{table_coefficients1} and Fig.~\ref{fig_bargraph_coeff_sparse} show
the variation of learned polynomial coefficients
($\bm{\alpha}=[\alpha_0,...,\alpha_9]$) for the 3rd-order system corresponding
to different $q$ values. These results show convergence to sparse solution when
$q\rightarrow 1$.
\begin{table}[hhh]
\caption{Estimated coefficient values over variation of $q$ values for a
3rd-order full polynomial based on 3 data samples}\label{table_coefficients1}
\centering{\scriptsize
\begin{tabular}{|c|r|r|r|r|r|r|}
\hline
            &   $q=5$ &     $q=2$  &   $q=1.75$ &    $q=1.5$ &    $q=1.3$ &  $q=1.1$ \\
\hline
 $\alpha_0$ & -3.934 &    -4.726 &    -4.852 &    -4.956 &    -4.996 &    -5.000 \\
 $\alpha_1$ & 12.013 &    17.881 &    18.856 &    19.662 &    19.967 &    20.000 \\
 $\alpha_2$ & 12.013 &    17.881 &    18.856 &    19.662 &    19.967 &    20.000 \\
 $\alpha_3$ & 14.754 &     3.624 &     1.761 &     0.394 &     0.019 &{\bf0.000} \\
 $\alpha_4$ & 16.268 &     5.459 &     3.107 &     0.985 &     0.103 &{\bf0.000} \\
 $\alpha_5$ & 16.268 &     5.459 &     3.107 &     0.985 &     0.103 &{\bf0.000} \\
 $\alpha_6$ & 13.646 &     0.729 &     0.185 &     0.012 & {\bf0.000}&{\bf0.000} \\
 $\alpha_7$ & 13.646 &     0.729 &     0.185 &     0.012 & {\bf0.000}&{\bf0.000} \\
 $\alpha_8$ & 15.510 &     1.280 &     0.405 &     0.041 &     0.001 &{\bf0.000} \\
 $\alpha_9$ & 15.510 &     1.280 &     0.405 &     0.041 &     0.001 &{\bf0.000} \\
\hline
\end{tabular} }
\end{table}

\section{Experiments on Prostate Cancer Data}

\subsection{Linear model fitting}

The data for this experiment was adopted in \cite{Hastie1} which came from a
study in \cite{Stamey1}. The correlation between the level of prostate-specific
antigen and several clinical measures were studied in men who were to receive a
radical prostatectomy. There are eight input variables with one response
output. Among the total 97 samples, 67 samples are used for training and the
remaining 30 samples are used for testing.

The 8 input variables are standardized by a Matlab \emph{zscore} normalization
\eqref{eqn_zscore} and then transformed by \eqref{eqn_exponential_transform}
with empirically chosen $a=1$ and $\bm{b}=\bm{\mu}$ for each dimension to move
the data to the first quadrant. Following the example in \cite{Hastie1}, a
linear model is adopted in this experiment. In other words, including the
intercept term, we have 9 parameters to be estimated.

\begin{figure}[hhh]
  \begin{center}
  \epsfxsize=12cm
  \epsfysize=8cm
  \epsffile[52   198   546   602]{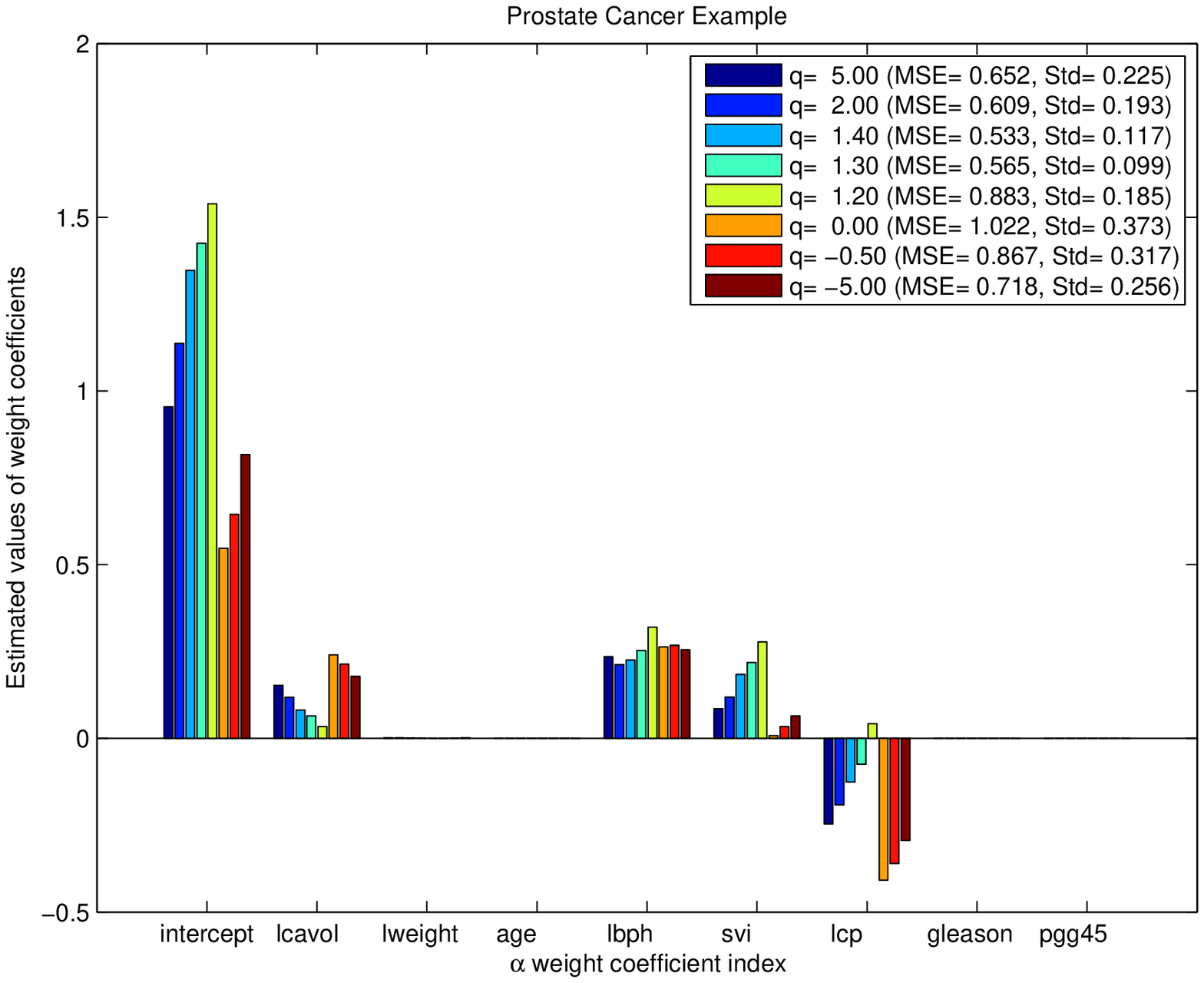}
  \caption{Stretchy regression: estimated coefficient values versus $q$ values
  ($q\in\{5,2,1.4,1.3,1.2,0,-0.5,-5\}$) for the prostate cancer data.}
  \label{fig_bargraph_coeff_q_exp}
  \end{center}
\end{figure}

\begin{figure}[hhh]
  \begin{center}
  \epsfxsize=12cm
  \epsfysize=8cm
  \epsffile[52   198   546   602]{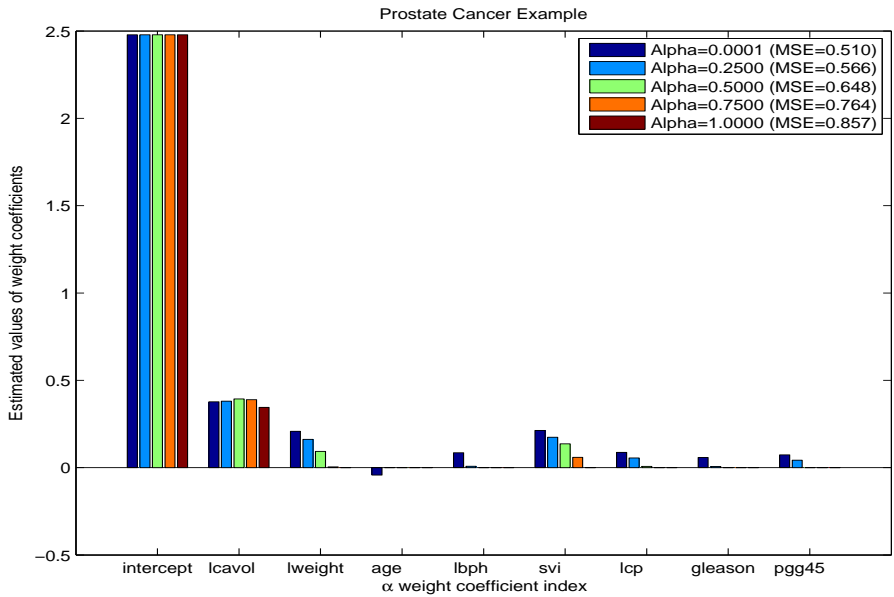}
  \caption{Lasso: estimated coefficient values versus \texttt{Alpha} values (\texttt{Alpha}$\in\{0.0001,0.25,0.5,0.75,1\}$)
  at \texttt{Lambda}=0.5 for the prostate cancer data.}
  \label{fig_bargraph_coeff_lasso}
  \end{center}
\end{figure}

The estimated weight parameters for the proposed stretchy regression for each
input variable including an intercept are shown in
Fig.~\ref{fig_bargraph_coeff_q_exp} and
Table~\ref{table_coefficients_prostate_q_exp} for
$q\in\{5,2,1.4,1.3,1.2,0,-0.5,-5\}$. Here we see stretching beyond the positive
and negative $q$ values can be feasible with reasonable accuracy. These results
show comparable test accuracy with that of lasso \cite{Tibshirani1} and
elastic-net \cite{Zou1}. In terms of model parameters, the results of
lasso-elastic-net from the statistical package of \cite{Matlab} as shown in
Fig.~\ref{fig_bargraph_coeff_lasso} and
Table~\ref{table_coefficients_prostate_lasso} show better convergence to
sparsity.

\begin{table}[hhh]
\caption{Stretchy regression: estimated coefficient values over variation of
$q$ values for prostate cancer data}\label{table_coefficients_prostate_q_exp}
\centering{\scriptsize
\begin{tabular}{|l|r|r|r|r|r|r|r|r|}
\hline
  parameter $\setminus$ $q$ &      $5$ &      $2$ &  $1.4$ &    $1.3$ &    $1.2$ &      $0$ &   $-0.5$ & $-5$\\
\hline
 $\alpha_0$: intercept&     0.954 &      1.137 &      1.347 &      1.425 &      1.539 &      0.547 &      0.645 &      0.817 \\
 $\alpha_1$: lcavol   &     0.153 &      0.119 &      0.082 &      0.065 &      0.034 &      0.240 &      0.214 &      0.179 \\
 $\alpha_2$: lweight  &     0.002 &      0.002 &      0.001 &      0.001 & {\bf 0.000}& {\bf-0.000}&      0.001 &      0.002 \\
 $\alpha_3$: age      &{\bf-0.000}& {\bf-0.000}& {\bf-0.000}& {\bf-0.000}& {\bf-0.000}& {\bf 0.000}& {\bf 0.000}& {\bf-0.000}\\
 $\alpha_4$: lbph     &     0.236 &      0.213 &      0.226 &      0.253 &      0.320 &      0.263 &      0.268 &      0.255 \\
 $\alpha_5$: svi      &     0.085 &      0.119 &      0.184 &      0.218 &      0.277 &      0.008 &      0.034 &      0.065 \\
 $\alpha_6$: lcp      &    -0.246 &     -0.191 &     -0.125 &     -0.074 &      0.042 &     -0.408 &     -0.360 &     -0.294 \\
 $\alpha_7$: gleason  &{\bf-0.000}& {\bf-0.000}& {\bf 0.000}& {\bf 0.000}& {\bf 0.000}& {\bf-0.000}& {\bf-0.000}& {\bf-0.000}\\
 $\alpha_8$: pgg45    &{\bf 0.000}& {\bf 0.000}& {\bf-0.000}& {\bf-0.000}& {\bf-0.000}& {\bf 0.000}& {\bf 0.000}& {\bf 0.000}\\
 \hline
 MSE                  &     0.652 &      0.609 &      0.533 &      0.565 &      0.883 &      1.022 &      0.867 &      0.718 \\
 STD                  &     0.225 &      0.193 &      0.117 &      0.099 &      0.185 &      0.373 &      0.317 &      0.256 \\
 \hline
\end{tabular} }
\end{table}

\begin{table}[hhh]
\caption{Lasso: estimated coefficient values over variation of \texttt{Alpha}
values at \texttt{Lambda}=0.5 for prostate cancer
data}\label{table_coefficients_prostate_lasso} \centering{\scriptsize
\begin{tabular}{|l|r|r|r|r|r|}
\hline
  parameter $\setminus$ \verb|Alpha|  & $0.0001$ &   $0.25$ &     $0.50$ &    $0.75$ &   $1.00$\\
\hline
  $\alpha_0$: intercept&    2.478 &      2.478 &      2.478 &      2.478 &      2.478 \\
  $\alpha_1$: lcavol   &    0.377 &      0.381 &      0.393 &      0.389 &      0.345 \\
  $\alpha_2$: lweight  &    0.208 &      0.162 &      0.093 &      0.004 &  {\bf0.000}\\
  $\alpha_3$: age      &   -0.042 &  {\bf0.000}&  {\bf0.000}&  {\bf0.000}&  {\bf0.000}\\
  $\alpha_4$: lbph     &    0.086 &      0.008 &  {\bf0.000}&  {\bf0.000}&  {\bf0.000}\\
  $\alpha_5$: svi      &    0.213 &      0.174 &      0.137 &      0.059 &  {\bf0.000}\\
  $\alpha_6$: lcp      &    0.087 &      0.056 &      0.007 &  {\bf0.000}&  {\bf0.000}\\
  $\alpha_7$: gleason  &    0.058 &      0.007 &  {\bf0.000}&  {\bf0.000}&  {\bf0.000}\\
  $\alpha_8$: pgg45    &    0.073 &      0.043 &  {\bf0.000}&  {\bf0.000}&  {\bf0.000}\\
  \hline
  MSE                  &    0.510 &      0.566 &      0.648 &      0.764 &      0.857 \\
\hline
\end{tabular} }
\end{table}

\subsection{Full polynomial model fitting}

In this experiment, we test the proposed stretchy regression with high order
polynomial models. Due to the large number of high order polynomial product
terms available for fitting, the inputs need appropriate scaling. We
empirically found that $a=10^{-5}$ and $\bm{b}=a\bm{\mu}$ for each dimension in
\eqref{eqn_exponential_transform} and $q=1.0001$ provides reasonable
performance. The small $a$ value is to scale the large summation of polynomial
product terms while a $q$ value relatively `close' to 1 can stretch the
suppression of parameters. To improve the stability of taking inverse of such a
large matrix, a regularization term of $10^{-4}$ is included in the matrix
inverse term during estimation. A stretchy dual ridge regression is performed
since the system is over-complete.

\begin{figure}[hhh]
  \begin{center}
  \epsfxsize=12cm
  \epsfysize=5cm
  \epsffile[60   198   550   616]{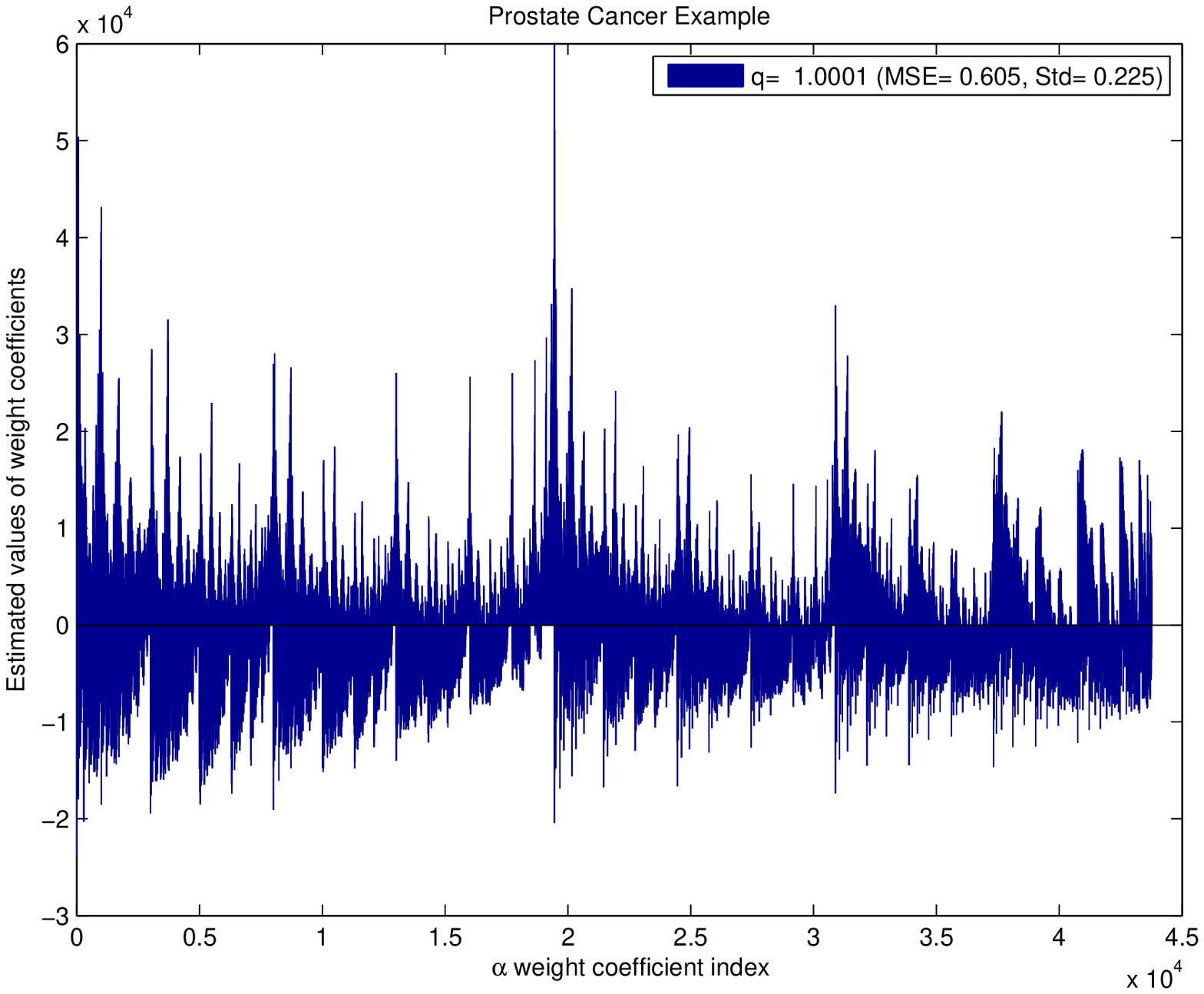}
  \caption{10th-order full polynomials: estimated coefficient values versus parameter index for the prostate cancer data.}
  \label{fig_bargraph_coeff_poly_10order}
  \end{center}
\end{figure}

Table~\ref{table_coefficients_numbers_MSE} shows the MSE results tabulated over
polynomial order settings. The corresponding number of polynomial expansion
terms are also tabulated along with the polynomial orders. These results show
feasibility of using a high order polynomials on high dimensional data when the
computational facility is suffice. Fig.~\ref{fig_bargraph_coeff_poly_10order}
shows the estimated 43758 parameters of the 10th order polynomial model. This
example shows the feasibility of high polynomial order adoption. However, an
in-depth study is yet desired for practical use since the estimated parameters
are of high magnitudes.

\begin{table}[hhh]
\caption{Estimation MSE with respect to polynomial order $r$ and number of
expansion terms $D$}\label{table_coefficients_numbers_MSE}
\centering{\scriptsize
\begin{tabular}{|l|r|r|r|r|r|r|r|r|r|r|}
\hline
 $r$ & 1     & 2     & 3     & 4     & 5     & 6     & 7     & 8     & 9     & 10    \\
\hline
 $D$ & 9     & 45    & 165   & 495   & 1287  & 3003  & 6435  & 12870 & 24310 & 43758 \\
\hline
 MSE & 0.738 & 0.497 & 0.518 & 0.537 & 0.549 & 0.560 & 0.571 & 0.584 & 0.596 & 0.605 \\
\hline
\end{tabular} }
\end{table}

\section{Conclusion}

A stretchy regression adopting a full multivariate polynomial model was
proposed in this article. Essentially, a warped closed-form solution was
derived in primal and dual forms analogous to that of ridge regression. Since
the solution operated upon positive real input values, an exponential
transformation was proposed to convert the inputs to the first quadrant of real
axes. Our preliminary experiments show effectiveness of the proposed method in
terms of compressive regression.

\section*{Acknowledgment}

The author is grateful to Dr. Geok-Choo Tan from NTU, Singapore for her
derivation of the number of full polynomial expansion terms presented in the
preliminary section.


\end{document}